\documentclass[11pt,a4paper]{article}
\usepackage[final]{nips_2017}

\usepackage[utf8]{inputenc} 
\usepackage[T1]{fontenc}    
\usepackage{hyperref}
\hypersetup{colorlinks,
citecolor=black
}      
\usepackage{url}            
\usepackage{booktabs}       
\usepackage{amsfonts}       
\usepackage{nicefrac}       
\usepackage{microtype}      
\usepackage{color}
\usepackage{graphicx}
\usepackage{amsmath}
\usepackage{subcaption}

\title{Mining Duplicate Questions of Stack Overflow}

\author{
  Mihir Sanjay Kale \\
  \texttt{mihirsak@andrew.cmu.edu} \\
  \and
  \textbf{Anirudha Rayasam}\\
  \texttt{arayasam@andrew.cmu.edu} \\
  \AND
  \textbf{Radhika Parik}\\
  \texttt{rparik@andrew.cmu.edu} \\
  \and
  \textbf{Pranav Dheram}\\
  \texttt{pdheram@andrew.cmu.edu} \\
}

\begin{document}

\maketitle

\begin{abstract}
There has a been a significant rise in the use of Community Question Answering sites (CQAs) over the last decade owing primarily to their ability to leverage the wisdom of the crowd. Duplicate questions have a crippling effect on the quality of these sites. Tackling duplicate questions is therefore an important step towards improving quality of CQAs. In this regard, we propose two neural network based architectures for duplicate question detection on Stack Overflow. We also propose explicitly modeling the code present in questions to achieve results that surpass the state of the art.
\end{abstract}

\section{Introduction}

There has a been a significant rise in the use of Community Question Answering sites (CQAs) over the last decade owing primarily to their ability to leverage a crowd’s collective intelligence. As CQAs continue to serve an increasing number of people every year, they also gain from expanding audience to become stronger in terms of the number of topics they cover. CQAs provide a platform where users can ask questions about a wide range of topics and receive replies from their peers who are better versed with these topics. Unfortunately, the increasing audience base has also made it hard for these sites to keep track of the quality of their content. Deterioration of quality of questions and answers has become more observable. Duplicate questions have a crippling effect on the quality of these sites. They increase the number of irrelevant search results forcing users to search longer. They also deter users from answering questions. Tackling duplicate questions is therefore an important step towards improving quality of CQAs. 

Consequently, there has been much research in the `duplicate question detection' \cite{muthmann2014automatic}\cite{ahasanuzzaman2016mining}\cite{bogdanova2015detecting} domain. Much of the work so far has used the text content of questions to build a predictive model for detecting duplicates. However, to the best of our knowledge, there has been very little work analysing the code snippets in these questions to identify duplicates. We believe that, in addition to using text content, we can leverage the large number of code snippets available on sites like stackoverflow to detect duplicates. We also present 2 architectures to effectively couple data from text and data from code to predict duplicate questions. 

The remainder of the paper is organized as follows. Section 2 briefly discusses previous work serving as a motivation for our own work. Section 3 motivates the use of code as a feature. Section 4 presents two baselines we compare our model with. Section 5 details our proposed approach and architecture. Section 6 discusses metrics to evaluate the proposed approach and Section 7 details the proposed timeline for carrying out this work and the conclusion is presented in Section 8. 

\section{Related Work}

\subsection{Predicting Q\&A Quality}
Shah et al. \cite{shah2010evaluating} propose classification techniques to identify metrics that predict answer quality on CQA sites. The study uses data from Yahoo Answers and builds classifiers based on crowd-sourced reviews of the answers as well as text-based and context-based features such as user profile, question length, number of answers etc extracted from the CQA website. The latter approach was more promising in terms of predicting the asker’s satisfaction with a given answer, than using human reviews on novelty, relevance etc. (which were found to be highly correlated). In the context of detecting duplicate questions, robust techniques to gauge answer quality are useful in ranking possible duplicate answers.  The results of Shah et al also served as a motivation to explore features derived from question metadata and text, rather than the realm of human subjectivity. 

Ravi et al. \cite{ravi2014great} predict question quality using latent topic models. Their work goes beyond the traditional bag of words model by leveraging latent structural clarity in questions. They use Latent Dirichlet Allocation(LDA) to generate latent topics which are used to build a global topic model(GTM). Better latent features can be extracted using LSTMs by leveraging time dependencies in text which we’ll discuss in more detail in our approach. And \cite{rajagopal2019domain} leverages semantic role labels as features in a transfer learning setting for low resource domains.

In AmazonQA \cite{gupta2019amazonqa} the authors develop a new dataset from the e-commerce product reviews and sets up the premise for a new reading comprehension task. They experiment with several models to benchmark the performance on the new dataset, including language model and span based QA models. Genetic algorithm based features selection \cite{anirudha2014genetic, anirudhaonline, anirudhasoft, anirudhaSCHPM} is another useful approach to identify the important features among a large set of features available.

\subsection{Graphical Analysis of CQA Networks}
Wang et al. \cite{wang2013wisdom} study the popular CQA platform Quora and attempt to identify factors behind its sustained growth and popularity. The authors compare Quora with Stack Overflow in terms distribution of content and activity, concluding that the platforms are quite similar. This observation renders the paper’s findings applicable to the Stack Overflow CQA ecosystem as well. The paper analyses three networks (graphs) from the Quora platform - user-topic Graph, social graph and question graph. User topic interaction is identified as an important aspect in determining user and question popularity, as is the quality and amount of content a user contributed to the site. The paper also explores graph clustering techniques to detect similar questions, an approach easily extensible to the detection of duplicate questions. The authors conclude that CQA websites face the challenge of pushing relevant content to their user-base in the face of increased activity, and with it, increased spam. This observation serves as an added motivation to our proposed research for detecting duplicate questions, as it is seen that question duplication is the most common reason behind question deletion on Stack Overflow. 

\subsection{Duplicate Question Detection}
There is a growing body of work on detecting duplicate questions in social media QA. Ahasanuzzaman et al. \cite{ahasanuzzaman2016mining} specifically look at detecting duplicate questions on the Stack Overflow platform. Their preliminary analysis establishes sufficient motivation for detecting and removing duplicate questions in the context of improving user experience on CQA sites as well as maintaining the website. The paper proposes to detect duplicates by training a discriminative classifier on duplicate and non-duplicate question pairs, using text-based similarity features extracted from question title, body, tags etc. Source code is explicitly mentioned as a feature with strong discriminative power, but is modeled only in terms of text similarity. This aspect of the paper motivated us to further explore the use of source code snippets in Stack Overflow to detect duplicate questions, especially when the underlying question similarity is indiscernible simply by looking at other features such as question title, question tags etc.

\section{Using Code as a Feature}

\begin{figure}[h]
  \begin{subfigure}[b]{0.4\textwidth}
  	\centering
    \includegraphics[width=\textwidth]{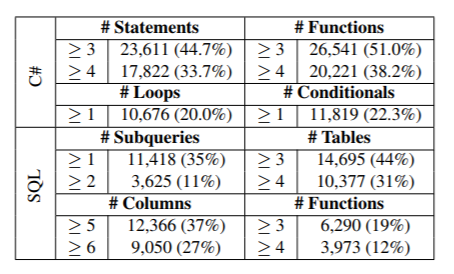}
    \caption{\centering Observations on Stack Overflow C\# and SQL code snippets \cite{iyer2016summarizing}}
    \label{code_snippet}
  \end{subfigure}
  \begin{subfigure}[b]{0.6\textwidth}
  \centering
    \includegraphics[width=\textwidth]{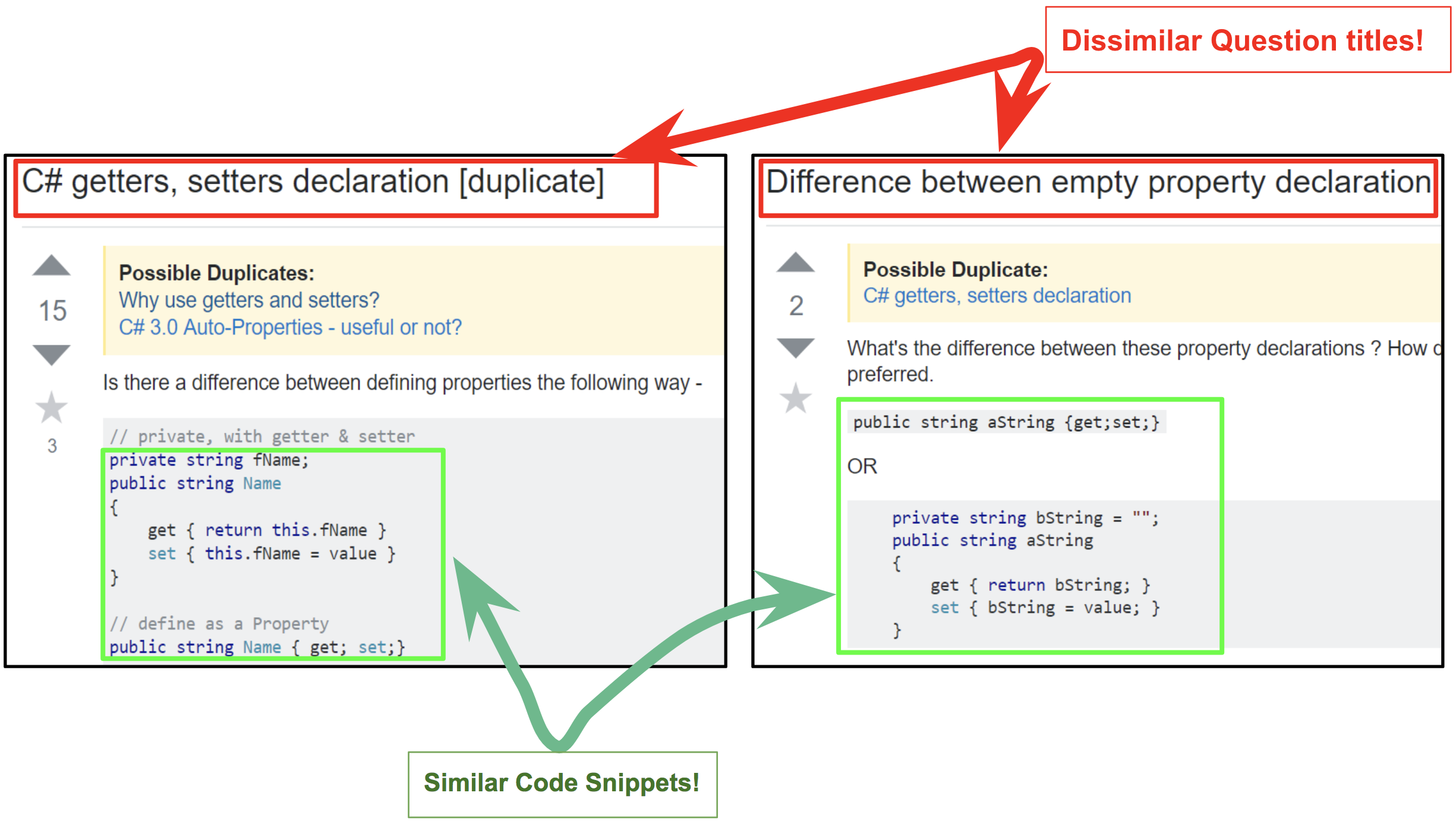}
    \caption{\centering Example of code being semantically a duplicate but syntactically different}
      \label{code_mot}
  \end{subfigure}
  \caption{Empirical motivation for code as a feature}
\end{figure}

Most research in the CQA domain uses features extracted from user profiles or metadata about the questions and answers \cite{shah2010evaluating} to classify questions. Even when the question text is used \cite{ahasanuzzaman2016mining, larionov2018tartan}, simple bag of words models are deployed which do not successfully leverage domain knowledge latent in the text. The research gap in using domain specific information(equations for math queries, code in tech questions) motivated us to use code snippets.Existing academic work in source code summarization by Iyer et al. \cite{iyer2016summarizing} attempts to summarize questions with code snippets (for C\# and SQL) on Stack Overflow. The paper concludes that the code snippets are usually representative of actual code\ref{code_snippet}.

The hypothesis that code snippets from CQA  questions add information other aspects of the question don’t capture may not seem intuitive at first. To highlight the discriminative power of code snippets in detecting duplicate questions, we present a motivating example from Stack Overflow \ref{code_mot} . The two questions \href{https://stackoverflow.com/questions/4923630/c-sharp-getters-setters-declaration}{stackexq1} and \href{https://stackoverflow.com/questions/7644542/difference-between-empty-property-declaration}{stackexq2} have been marked as duplicates by Stack Overflow moderators, but the overlap in question title and text would easily elude an ordinary text based classifier. A closer look at the source code structure and semantics reveals that the questions are in fact quite similar. Intuitively, this follows from the fact that coding standards and idiomatic code renders code snippets more similar than natural language text, despite differences in individual coding style.

\section{Baselines}
Following from the previous section it is very clear that including code for duplicate detection is of importance. In this section we brief about some of the prior work which incorporates code as a feature for duplicate question detection in social media question and answering forums and present their effectiveness and shortcomings. 

\subsection{CNN based Siamese Network}
Bogdanova et al \cite{bogdanova2015detecting} use a Siamese network inspired CNN based duplication detection network. A CNN is trained on the word2vec embeddings of the questions with the objective of maximizing the cosine distance between duplicate questions and vice versa. They achieve an accuracy of 92.9\% on the duplicate detection task on the AskUbuntu and Meta Stack Exchange dataset.

However, this method has several limitations. It’s hard to model long term dependencies often found in language and code using convolutional filters, since filters are better suited for local context. Moreover, embeddings are learnt on title + body + code together. But code has a very different structure and semantics than English. Code is not made a first class citizen in learning a representation for the question.

\subsection{Mining Duplicate Questions in Stack Overflow}
In the work by Ahasanuzzaman et al. \cite{ahasanuzzaman2016mining} which uses the MSR 2015 Stack Overflow data dump, they train a Logistic Regression based binary classifier model to determine if a pair of input questions are duplicate or not. The model is trained with input as a pair of questions, which is obtained by generating all possible question pairing and retaining only an equal number of positive(duplicate) and negative pairs to avoid class imbalance, and the binary output is parsed from the dataset. The question content excluding the code is preprocessed using Porter Stemming and NER, and the embeddings are obtained using WordNet Similarity synsets and represented as a bag-of-embeddings. The code is represented as an annotated bag-of-words, with annotations to differentiate the code keywords from the text, and concatenated with the text representation. The similarity between the questions are computed based on the cosine similarity between the corresponding representations. 

The highlight of the work is the increase in recall@5 from 45.8 to 51.2 after incorporating code related features, illustrating the effectiveness of code. Despite this increase the method is still plagued with limitations as n-grams with a linear model may not be expressive enough to model text, let alone modeling code, both of which may have several long-term structural and semantic dependencies. This is a very strong motivation to exploit Long Short-Term Memory(LSTMs) to capture these dependencies and hence further reinforces our proposal decision.

\section{Proposed Approach}

This section describes the approach we adopt for the detection of duplicate questions. A brief description of the dataset being used and the construction of training and test set is provided, followed by detailing of our LSTM model incorporated for learning representations for text and code features. Finally, we present two different approaches we want to experiment with during the course of the project. The first model is inspired by a Siamese network and the second is a CNN based approach which learns over a feature-grid based representation of the question representations. 

\subsection{Training-Dataset Construction}
The proposed approach requires data for two tasks:
\\

\textbf{1. Training the Language Model} - We crawl GitHub to extract code that performs the same task in order to obtain code similar in functionality but different in style. This dataset can be augmented to a language agnostic model using tools such as Java2CSharp.
\\

\textbf{2. Training the Classifier} - We extract pairs of similar and dissimilar questions containing code snippets from the StackExchange Data Dump (a Stack Overflow dataset containing question text and associated metadata) [\href{https://archive.org/details/stackexchange}{stackex}]

Duplicate question (positive) pairs are already available in the Stack Overflow dataset. We sample questions from the data to generate an equal number of non-duplicate (negative) pairs. Random sampling will create pairs that are  semantically very different, making it easy for the classifier to discriminate just based on keywords. In the production scenario, however, it is more important to be able to correctly classify questions that are semantically closer. Therefore to make the model more robust we create negative pairs by matching non-duplicate questions that are similar to each other. 

For instance a question about the pandas library in python will be matched with a another pandas library question. We note that there is a chance that negative samples may in fact be positive since we do not have at our disposal the entire universe of duplicates.
However the probability of this happening is very small since the fraction of questions that are duplicates is small. Thus, the degree of impurity in our dataset will be very low.

Finally, we will use 80\% of the data for training, 10\% for validation and 10\% for testing.

\subsection{LSTM based feature learning}
LSTMs have recently shown great results for a variety of tasks in NLP, including
machine translation, named entity recognition etc. The overall architecture of
an LSTM is shown in Figure 1. They have the ability to encode long term dependencies present in sequence data. Visualizations by Karpathy et al \cite{karpathy2015visualizing} show that an LSTM can learn associations like `\{' is followed by `\}' even though they may be separated by several lines of code. In the following subsections we detail the methodology to learn the representations for text and code content of a question.

\begin{figure}
	\centering
	\includegraphics[width = 0.85\textwidth]{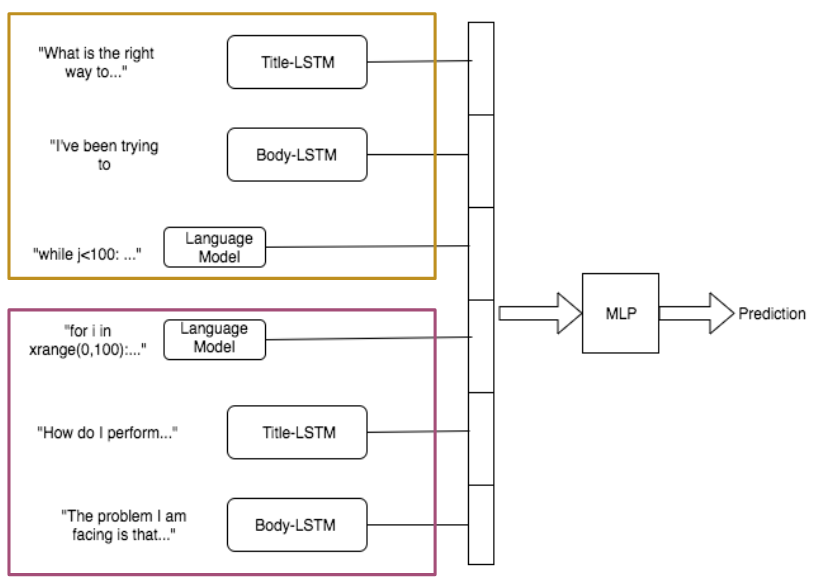}
    \caption{LSTM based architecture}\label{fig:lstm}
\end{figure}

\subsubsection{Representation Learning for Code Snippets}
To efficiently exploit information present in code, we need to find a way to represent a code snippet. We do not have enough code in the Stack Overflow dataset to learn a good representation. Secondly, the number of paired duplicate code snippets is even lower, making it hard to learn a supervised embedding. Therefore, we look at unsupervised methods to leverage the large amount of source code available online via services such as Github.

\subsubsection{Language Model for Code}
First, we create language specific corpora (separate corpus for Python, Java etc) by crawling Github. The crawled dataset of language specific code will be made public. 
First, the variable and class names in code snippets are all standardized. Such names can be different across different pieces of code. For instance, if a program has variable names - count, a, b, temp we will standardize them to var1, var2,var3,var4. After this the snippet is tokenized and rare tokens are removed.
On a given corpus, we learn a character level LSTM which aims to predict the next character of a sequence []. At the end of the training process, we have an LSTM which acts as a language model for the given programming language. For a code snippet, we run it through our LSTM and use the hidden representation at the last times step as the embedding for the snippet. This embedding encodes the  syntactic and semantic information present in the code and can be used as input to our main model which identifies duplicate questions. We train a different model for each language (Python, Java etc).

\subsection{Model Architecture}

\subsubsection{LSTM Based Siamese Network}
In this section we describe our overall model architecture. A question mainly consists of three parts - title , body and code as can be seen in Figure \ref{fig:lstm}. Each part contains information presented in a different way. The title is concise and generally 1-2 sentences long and summarizes the question. The body is longer, more detailed and contains all the relevant technical information regarding the problem the user is facing. The code , if present, give the most direct indication of the errors faced by the asker. Since each section has a different intention and different semantics,  we encode them separately. Our model takes two questions as input and returns a probabilistic score between 0 and 1, where 1 implies that the questions are duplicates.

Each question is split into body, title and code. The code embedding is described in the previous section. The title and body embeddings are learnt via LSTMs as seen in Figure \ref{fig:lstm}.The weights of the two Title-LSTMs in are shared. The weights of the Body-LSTMs are also shared. 
In the next layer, we concatenate the all the embeddings from the two questions. A MLP is used on top of this layer to learn higher order interactions between the embeddings in order to predict duplicates. The MLP has a sigmoid layer to output the probabilistic score. While the code embeddings are from our unsupervised language model, the embeddings for Title and LSTM are learnt from scratch via supervision provided by the duplicate labels.

\begin{figure}
	\centering
	\includegraphics[width = 0.6\textwidth]{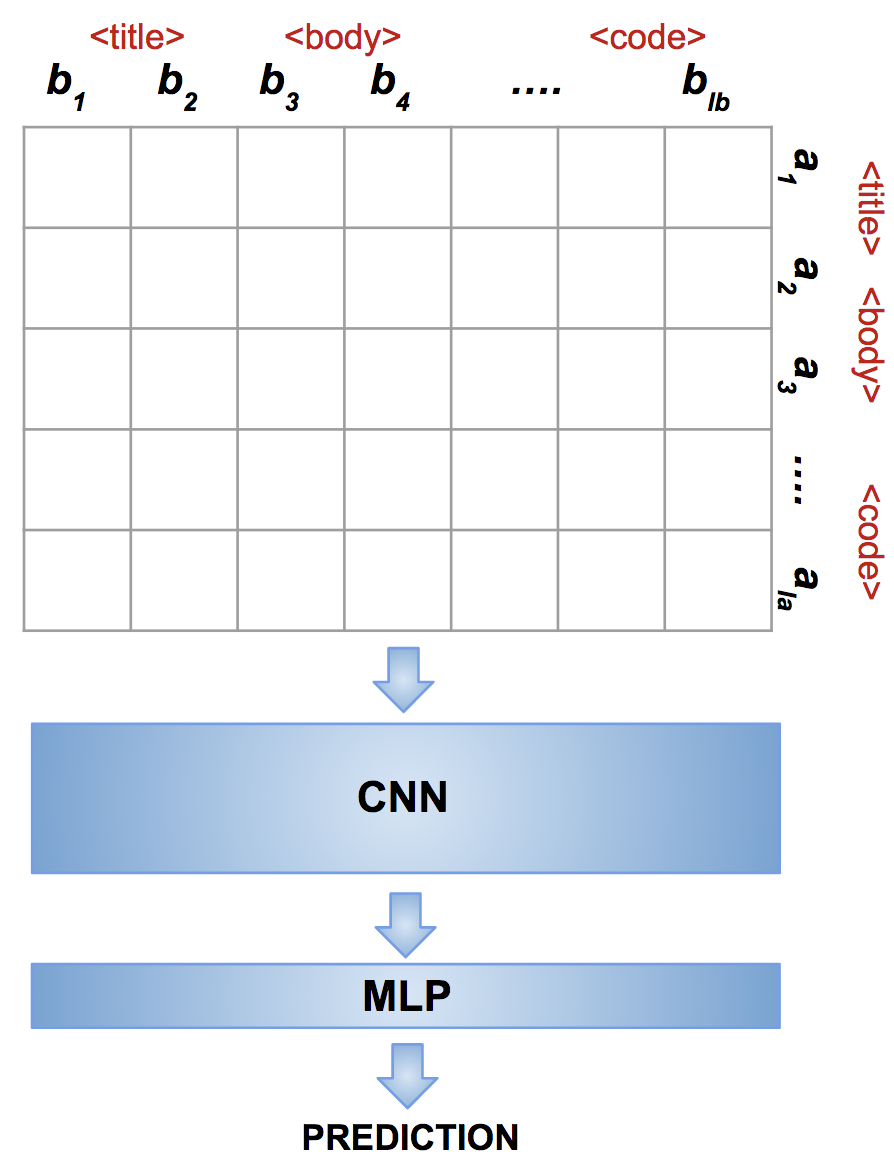}
    \caption{Architecture for Feature-Grid based CNN Model}\label{feature-grid}
\end{figure}

\subsubsection{Feature-grid Based CNN Approach}
The previous proposed model and also most other prior models are not capable of automatically handling alignment of question content, as the details in each question can be structurally differently. Also, the previous model incorporates the code features only when both the questions in the pair have code in them. These subtleties might result in information loss during computation of similarity between vector representations as they are linearly represented. Finally, most of the computations involved are very intensive and require high parallelization lest the training becomes very time consuming. In order to counter these shortcomings we propose representing the question features as a grid and attempt to learn the similarity patterns by training a Convolutional Neural Networks(CNN) based model over it.

Given two questions \textit{a} and \textit{b}, we learn the LSTM based embedding representations for the text and code as explained in the previous sections and then for each question we concatenate the representations for title, body and code computed independently to obtain joint embedding vectors of length $l_a$ and $l_b$ for question \textit{a} and \textit{b} respectively. The final representation of each question is as follows: 

\begin{align}
a &= (a_1, a_2, ... , a_i , ... a_{l_a}) \nonumber \\
\nonumber \\
b &= (b_1, b_2, ... , b_i , ... b_{l_b}) \nonumber
\end{align}

where each $a_i,b_j \in R^d$, is a word embedding vector of dimension \textit{d}. The feature-grid, as seen in Figure \ref{feature-grid}, constructed from the obtained embedding vectors is of $l_a \times l_b$ dimension and the value at any position [\textit{i,j}] in the grid is the similarity score between embeddings $a_i$ and $b_j$. We train a CNN based classification model over this feature grid to learn to classify whether given questions are duplicate or not. In short, the input from the feature-grid is fed into the CNN layer whose output is passed through a Multi-Layer Perceptron to get a binary output indicating the duplicity. 

\section{Evaluation}
Since we have cast our problem in the binary classification setting, we use standard classification metrics such as accuracy, precision, recall, auROC for evaluation. These metrics also help us directly compare our models with current state of the art.

\section{Conclusion}

It has become increasingly difficult to control the nuisance of duplicate questions in QA platforms, due to the unprecedented growth in popularity of these platforms as well as the shortage if human curators. This is detrimental to query search results and overall user experience and hence we proposed a model for question duplicate detection. The novelty factor in our proposed approach to duplication detection is using a distributional representation of source code as a feature in out model. Using code embeddings as inputs to an LSTM model would enable the detection of semantic and logical similarity across seemingly different code snippets. We also propose two neural network based architectures for duplicate question detection on Stack Overflow.

\bibliographystyle{apalike}
\bibliography{mltm}
\end{document}